\documentclass[10pt,twocolumn,letterpaper]{article}

\usepackage{iccv}
\usepackage{times}
\usepackage{epsfig}
\usepackage{graphicx}
\usepackage{amsmath}
\usepackage{amssymb}
\usepackage[shortlabels]{enumitem}
\usepackage[normalem]{ulem}
\usepackage{multirow}
\usepackage{threeparttable}
\usepackage[tight,normalsize,sf,SF]{subfigure}

\usepackage[pagebackref=true,breaklinks=true,letterpaper=true,colorlinks,bookmarks=false]{hyperref}

\newcommand{\kate}[1]{\textcolor{magenta}{}}
\newcommand{\huijuan}[1]{\textcolor{green}{}}

\newcommand{\modelnamelong}[0]{\textit{Region Convolutional 3D Network (R-C3D) }}
\newcommand{\modelname}[0]{R-C3D }

\iccvfinalcopy 

\def\iccvPaperID{2582} 

\ificcvfinal\fi

\begin{document}

\title{R-C3D: Region Convolutional 3D Network for Temporal Activity Detection}


\author{Huijuan Xu~~~~~~~~~~~~~~~~~~~~~~~~
Abir Das~~~~~~~~~~~~~~~~~~~~~~~~
Kate Saenko\\
Boston University\\
Boston, MA\\
{\tt\small \{hxu, dasabir, saenko\}@bu.edu}
}

\maketitle

\begin{abstract} 
We address the problem of activity detection in continuous, untrimmed video streams.
This is a difficult task that requires extracting meaningful spatio-temporal features to capture activities, accurately localizing the start and end times of each activity.
We introduce a new model, \textit{Region Convolutional 3D Network (R-C3D)}, which encodes the video streams using a three-dimensional fully convolutional network, then generates candidate temporal regions containing activities, and finally classifies selected regions into specific activities.
Computation is saved due to the sharing of convolutional features between the proposal and the classification pipelines.
The entire model is trained end-to-end with jointly optimized localization and classification losses.
\modelname is faster than existing methods (569 frames per second on a single Titan X Maxwell GPU) and achieves state-of-the-art results on THUMOS'14.
We further demonstrate that our model is a general activity detection framework that does not rely on assumptions about particular dataset properties by evaluating our approach on ActivityNet and Charades.
Our code is available at \url{http://ai.bu.edu/r-c3d/}

\end{abstract} 
\vspace{-0.2in}
\section{Introduction}

Activity detection in continuous videos is a challenging problem that requires not only recognizing, but also precisely localizing activities in time.
Existing state-of-the-art approaches address this task as \textit{detection by classification}, \textit{i.e.} classifying temporal segments generated in the form of sliding windows~\cite{karaman2014fast, oneata2014lear, shou2016temporal, wang2014action} or via an external ``proposal'' generation mechanism~\cite{caba2016fast, Wang2016}. These approaches suffer from one or more of the following major drawbacks: they do not learn deep representations in an end-to-end fashion, but rather use hand-crafted features~\cite{wang2013action, wang2014video}, or deep features like VGG~\cite{simonyan2014very}, ResNet~\cite{he2016deep}, C3D~\cite{tran2015learning} \textit{etc.}, learned separately on image/video classification tasks. Such off-the-shelf representations may not be optimal for localizing activities in diverse video domains, resulting in inferior performance. Furthermore, current methods' dependence on external proposal generation or exhaustive sliding windows leads to poor computational efficiency. Finally, the sliding-window models cannot easily predict flexible activity boundaries.

\begin{figure}[t]
\begin{center}
\includegraphics[width=0.75\linewidth]{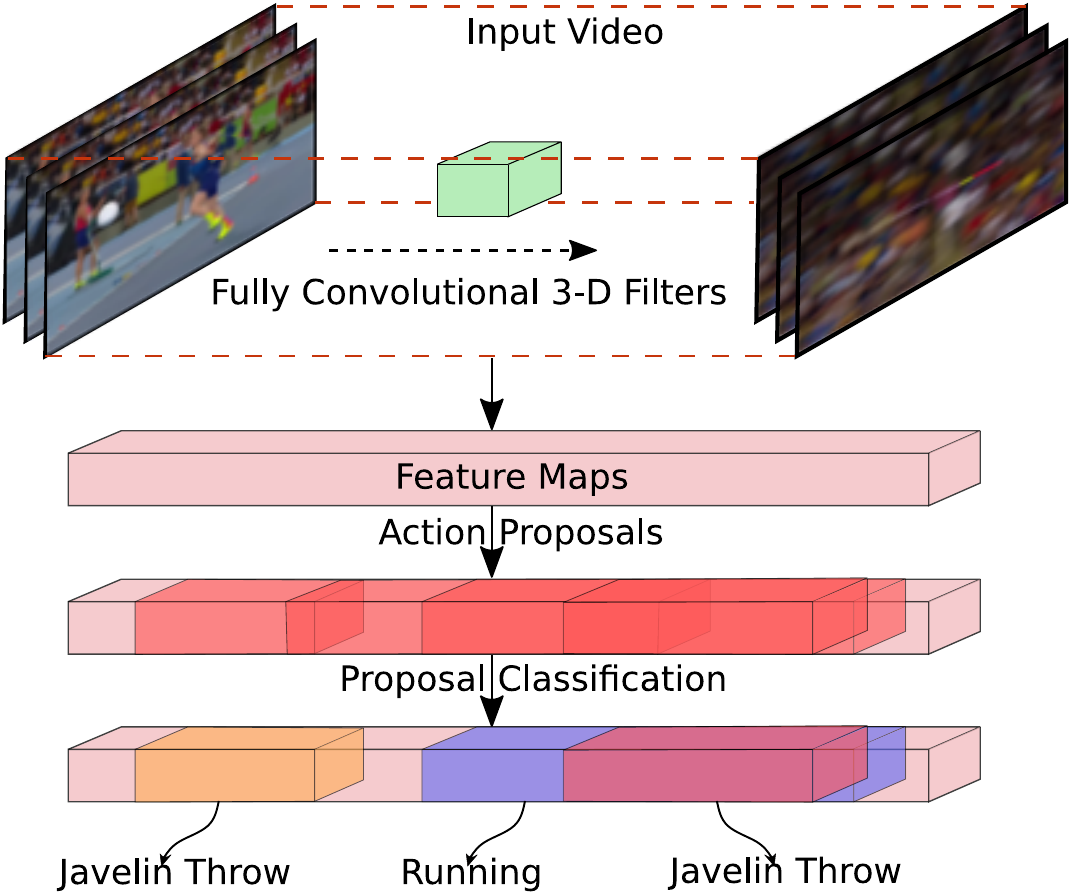}
\end{center}
\vskip -0.15in
\caption{We propose a fast end-to-end \modelnamelong for activity detection in continuous video streams. The network encodes the frames with fully-convolutional 3D filters, proposes activity segments, then classifies and refines them based on pooled features within their boundaries. Our model improves both speed and accuracy compared to existing methods.}
\label{fig:schematic}
\vskip -0.2in
\end{figure}

In this paper, we propose an activity detection model that addresses all of the above issues. Our \modelnamelong is end-to-end trainable and learns task-dependent convolutional features by jointly optimizing proposal generation and activity classification.
Inspired by the Faster R-CNN~\cite{ren2015faster} object detection approach, we compute fully-convolutional 3D ConvNet features and propose temporal regions likely to contain activities, then pool features within these 3D regions to predict activity classes (Figure~\ref{fig:schematic}).
The proposal generation stage filters out many background segments and results in superior computational efficiency compared to sliding window models.
Furthermore, proposals are predicted with respect to predefined anchor segments and can be of arbitrary length, allowing detection of flexible activity boundaries.

Convolutional Neural Network (CNN) features learned end-to-end have been  successfully used for activity recognition~\cite{karpathy2014large, simonyan2014two}, particularly in 3D ConvNets (C3D~\cite{tran2015learning}), which learn to capture spatio-temporal features.
However, unlike the traditional usage of 3D ConvNets~\cite{tran2015learning} where the input is short 16-frame video chunks, our method applies full convolution along the temporal dimension to encode as many frames as the GPU memory allows. Thus, rich spatio-temporal features are automatically learned from longer videos.
These feature maps are shared between the activity proposal  and classification subnets to save computation time and jointly optimize features for both tasks.

Alternative activity detection approaches~\cite{escorcia2016daps, ma2016learning, montes2016temporal, Singh2016a, yeung2016end} use a recurrent neural network (RNN) to encode a sequence of frame or  video chunk features (\textit{e.g.} VGG~\cite{simonyan2014very}, C3D~\cite{tran2015learning}) and predict the activity label at each time step. However, these RNN methods can only model temporal features at a fixed granularity (e.g. per-frame CNN features or 16-frame C3D features). 
In order to use the same classification network to classify variable length proposals into specific activities, we extend 2D region of interest (RoI) pooling to 3D which extracts a fixed-length feature representation for these proposals. Thus, our model can utilize video features at any temporal granularity.
Furthermore, some RNN-based detectors rely on direct regression to predict the temporal boundaries for each activity.
As shown in object detection~\cite{girshick2014rich, Szegedy2013} and semantic segmentation~\cite{Carreira2012}, object boundaries obtained using a regression-only framework are inferior compared to ``proposal based detection''.

We perform extensive comparisons of \modelname to state-of-the-art activity detection methods  using three publicly available benchmark datasets - THUMOS'14~\cite{THUMOS14}, ActivityNet~\cite{caba2015activitynet} and Charades~\cite{sigurdsson2016hollywood}.
We achieve new state-of-the-art results on THUMOS'14 and Charades, and  improved results on ActivityNet when using only C3D features.

To summarize, the main contributions of our paper are:
\begin{itemize}[noitemsep,nolistsep]
    \item an end-to-end activity detection model with combined activity proposal and classification stages that can detect arbitrary length activities;
    \item fast detection speeds (5x faster than current methods) achieved by sharing fully-convolutional C3D features between the proposal generation and classification parts of the network;
    \item extensive evaluations on three diverse activity detection datasets that demonstrate the general applicability of our model.
\end{itemize}

\section{Related Work}

\textbf{Activity Detection}
There is a long history of activity recognition, or classifying trimmed video clips into fixed set of categories~\cite{ji20133d, laptev2008learning, yue2015beyond, simonyan2014two, wang2013action, Zheng2016}.
Activity \textit{detection} also needs to predict the start and end times of the activities within untrimmed and long videos.
Existing activity detection approaches are dominated by models that use sliding windows to generate segments and subsequently classify them with activity classifiers trained on multiple features~\cite{karaman2014fast, oneata2014lear, shou2016temporal, wang2014action}.
Most of these methods have stage-wise pipelines which are not trained end-to-end.
Moreover, the use of exhaustive sliding windows is computationally inefficient and constrains the boundary of the detected activities to some extent.

Recently, some approaches have bypassed the need for exhaustive sliding window search to detect activities with arbitrary lengths.
\cite{escorcia2016daps, ma2016learning, montes2016temporal, Singh2016a, yeung2016end} achieve this by modeling the temporal evolution of activities using RNNs or LSTMs networks and predicting an activity label at each time step.
The deep action proposal model~\cite{escorcia2016daps} uses LSTM to encode C3D features of every 16-frame video chunk, and directly regresses and classifies activity segments without the extra proposal generation stage.
Compared to this work, we avoid recurrent layers, encoding a large video buffer with a fully-convolutional 3D ConvNet, and use 3D RoI pooling to allow feature extraction at arbitrary proposal granularity, achieving significantly higher accuracy and speed.
The method in~\cite{yuan2016temporal} tries to capture motion features at multiple resolutions by proposing a Pyramid of Score Distribution Features.
However their model is not end-to-end trainable and relies on handcrafted features.

Aside from supervised activity detection, a recent work~\cite{Wang2017UntrimmedNets} has addressed weakly supervised activity localization from data labeled only with video level class labels by learning attention weights on shot based or uniformly sampled proposals.
The framework proposed in~\cite{Richard_2016_CVPR} explores the uses of a language model and an activity length model for detection.
Spatio-temporal activity localization~\cite{Weinzaepfel_2015_ICCV, Yu_2015_CVPR} have also been explored to some extent.
We only focus on supervised temporal activity localization.

\begin{figure*}[ht]
\begin{center}
\includegraphics[width=\linewidth]{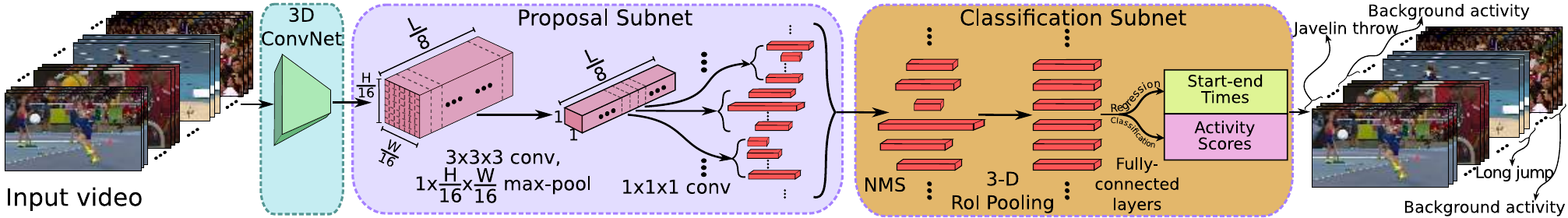}
\end{center}
\vskip -0.15in
\caption{\modelname model architecture.
The 3D ConvNet takes raw video frames as input and computes convolutional features.
These are input to the Proposal Subnet that proposes candidate activities of variable length along with confidence scores.
The Classification Subnet filters the proposals, pools fixed size features and then predicts activity labels along with refined segment boundaries.}
\label{fig:architecture}
\end{figure*}

\textbf{Object Detection}
Activity detection in untrimmed videos is closely related to object detection in images.
The inspiration for our work, Faster R-CNN~\cite{ren2015faster}, extends R-CNN~\cite{girshick2014rich} and Fast R-CNN~\cite{girshick2015fast} object detection approaches, incorporating RoI pooling and a region proposal network.
Compared to recent object detection models \textit{e.g.}, SSD~\cite{liu2016ssd} and R-FCN~\cite{li2016r}, Faster R-CNN is a general and robust object detection framework that has been deployed on different datasets with little data augmentation effort. 
Like Faster R-CNN, our \modelname model is also designed with the goal of easy deployment on varied activity detection datasets.
It avoids making certain assumptions based on unique characteristics of a dataset, such as the UPC model for ActivityNet~\cite{montes2016temporal} which assumes that each video contains a single activity class.
We show the effectiveness of our model on three different types of activity detection datasets, the most extensive evaluation to our knowledge.

\section{Approach}
\label{sec:approach}

We propose a \textit{Region Convolutional 3D Network (R-C3D)}, a novel convolutional neural network for activity detection in continuous video streams.
The network, illustrated in Figure~\ref{fig:architecture}, consists of three components: a shared 3D ConvNet feature extractor~\cite{tran2015learning}, a temporal proposal stage, and an activity classification and refinement stage.
To enable efficient computation and end-to-end training, the proposal and classification sub-networks share the same C3D feature maps.
The proposal subnet predicts variable length temporal segments that potentially contain activities, while the classification subnet classifies these proposals into specific activity categories or background, and further refines the proposal segment boundaries.
A key innovation is to extend the 2D RoI pooling in Faster R-CNN to 3D RoI pooling which allows our model to extract features at various resolutions for variable length proposals.
Next, we describe the shared video feature hierarchies in Sec.~\ref{sec:feature}, the temporal proposal subnet in Sec.~\ref{sec:proposal} and the classification subnet in Sec.~\ref{sec:classification}. Sections~\ref{sec:optimization} and \ref{sec:prediction} detail the optimization strategy during training and testing respectively.

\subsection{3D Convolutional Feature Hierarchies}
\label{sec:feature}
We use a 3D ConvNet to extract rich spatio-temporal feature hierarchies from a given input video buffer.
It has been shown that both spatial and temporal features are important for representing videos, and a 3D ConvNet encodes rich spatial and temporal features in a hierarchical manner.
The input to our model is a sequence of RGB video frames with dimension $\mathbb{R}^{3\times L\times H\times W}$.
The architecture of the 3D ConvNet is taken from the C3D architecture proposed in~\cite{tran2015learning}.
However, unlike~\cite{tran2015learning}, the input to our model is of variable length.
We adopt the convolutional layers (\texttt{conv1a} to \texttt{conv5b}) of C3D, so a feature map $C_{conv5b}\in \mathbb{R}^{512\times \frac{L}{8} \times \frac{H}{16}\times \frac{W}{16}}$ ($512$ is the channel dimension of the layer \texttt{conv5b}) is produced as the output of this subnetwork.
We use $C_{conv5b}$ activations as the shared input to the proposal and classification subnets.
The height ($H$) and width ($W$) of the frames are taken as 112 each following~\cite{tran2015learning}.
The number of frames $L$ can be arbitrary and is only limited by memory.

\subsection{Temporal Proposal Subnet}
\label{sec:proposal}
To allow the model to predict variable length proposals, we incorporate anchor segments into the temporal proposal sub-network.
The subnet predicts potential proposal segments with respect to anchor segments and a binary label indicating whether the predicted proposal contains an activity or not.
The anchor segments are pre-defined multiscale windows centered at $L/8$ uniformly distributed temporal locations.
Each temporal location specifies $K$ anchor segments, each at a different fixed scale.
Thus, the total number of anchor segments is $(L/8) * K$.
The same set of $K$ anchor segments exists in different temporal locations, which ensures that the proposal prediction is temporally invariant.
The anchors serve as reference activity segments for proposals at each temporal location, where the maximum number of scales $K$ is dataset dependent.

To obtain features at each temporal location for predicting proposals with respect to these anchor segments, we first add a 3D convolutional filter with kernel size $3\!\!\times \!\!3\!\!\times \!\!3$ on top of $C_{conv5b}$ to extend the temporal receptive field for the temporal proposal subnet.
Then, we downsample the spatial dimensions (from $\frac{H}{16}\!\times \!\frac{W}{16}$ to $1\!\times \!1$) to produce a \textit{temporal} only feature map $C_{tpn}\in \mathbb{R}^{512\times \frac{L}{8} \times 1\times 1}$ by applying a 3D max-pooling filter with kernel size $1\!\times \!\frac{H}{16} \!\!\times \!\!\frac{W}{16}$.
The 512-dimensional feature vector at each temporal location in $C_{tpn}$ is used to predict a relative offset $\{\delta c_{i},\delta l_{i}\}$ to the center location and the length of each anchor segment $\{c_{i},l_{i}\},i\in \{1,\cdots,K\}$.
It also predicts the binary scores for each proposal being an activity or background.
The proposal offsets and scores are predicted by adding two $1\!\times \!1\!\times \!\!1$ convolutional layers on top of $C_{tpn}$.

\noindent{\textbf{Training}:} For training, we need to assign positive/negative labels to the anchor segments.
Following the standard practice in object detection~\cite{ren2015faster}, we choose a positive label if the anchor segment 1) overlaps with some ground-truth activity with Intersection-over-Union (IoU) higher than 0.7, or 2) has the highest IoU overlap with some ground-truth activity.
If the anchor has IoU overlap lower than 0.3 with all ground-truth activities, then it is given a negative label. All others are held out from training.
For proposal regression, ground truth activity segments are transformed with respect to nearby anchor segments using the coordinate transformations described in Sec.~\ref{sec:optimization}.
We sample balanced batches with a positive/negative ratio of $1\!:\!1$.

\subsection{Activity Classification Subnet}
\label{sec:classification}
The activity classification stage has three main functions: 1) selecting proposal segments from the previous stage, 2) three-dimensional region of interest (3D RoI) pooling to extract fixed-size features for selected proposals, and 3) activity classification and boundary regression for the selected proposals based on the pooled features. 

Some activity proposals generated by the proposal subnet highly overlap with each other and some have a low proposal score indicating low confidence.
Following the standard practice in object detection~\cite{Felzenszwalb2010Object, ren2015faster} and activity detection~\cite{shou2016temporal, yeung2016end}, we employ a greedy Non-Maximum Suppression (NMS) strategy to eliminate highly overlapping and low confidence proposals.
The NMS threshold is set as 0.7.

The selected proposals can be of arbitrary length.
However we need to extract fixed-size features for each of them in order to use fully connected layers for further activity classification and regression.
We design a 3D RoI pooling layer to extract the fixed-size volume features for each variable-length proposal from the shared convolutional features $C_{conv5b} \in \mathbb{R}^{512\times (L/8) \times 7\times 7}$ (shared with the temporal proposal subnet).
Specifically, in 3D RoI pooling, an input feature volume of size, say, $ l \! \times\! h \!\times\! w$ is divided into $ l_s \!\times \!h_s\!\times \!w_s$ sub-volumes each with approximate size $ \frac{l}{l_s} \!\times\! \frac{h}{h_s} \!\times\! \frac{w}{w_s}$, and then max pooling is performed inside each sub-volume.
In our case, suppose a proposal has the feature volume of $ l_p \!\times\! 7 \!\times\! 7$ in $C_{conv5b}$, then this feature volume will be divided into $1 \!\times\! 4 \!\times\! 4$ grids and max pooled inside each grid.
Thus, proposals of arbitrary lengths give rise to  output volume features of the same size $512 \!\times\! 1 \!\times\! 4 \!\times\! 4$.

The output of the 3D RoI pooling is fed to a series of two fully connected layers.
Here, the proposals are classified to activity categories by a classification layer and the refined start-end times for these proposals are given by a regression layer.
The classification and regression layers are also two separate fully connected layers and for both of them the input comes from the aforementioned fully connected layers (after the 3D RoI pooling layer).

\noindent\textbf{Training:} We need to assign an activity label to each proposal for training the classifier subnet.
An activity label is assigned if the proposal has the highest IoU overlap with a ground-truth activity, and at the same time, the IoU overlap is greater than 0.5.
A background label (no activity) is assigned to proposals with IoU overlap lower than 0.5 with all ground-truth activities.
Training batches are chosen with positive/negative ratio of $1\!:\!3$.


\subsection{Optimization}
\label{sec:optimization}
We train the network by optimizing both the classification and regression tasks jointly for the two subnets.
The softmax loss function is used for classification, and smooth L1 loss function~\cite{girshick2015fast} is used for regression.
Specifically, the objective function is given by:
\small
\begin{equation}
    \hspace{-2mm} Loss = \frac{1}{N_{cls}} \sum\limits_i L_{cls} (a_i, a_i^*) + \lambda \frac{1}{N_{reg}} \sum\limits_i a_i^* L_{reg} (t_i, t_i^*)
\vspace{-2mm}
\label{eq:loss}
\end{equation}
\normalsize
where $N_{cls}$ and $N_{reg}$ stand for batch size and the number of anchor/proposal segments, $\lambda$ is the loss trade-off parameter and is set to a value $1$.
$i$ is the anchor/proposal segments index in a batch, $a_i$ is the predicted probability of the proposal or activities, $a_i^*$ is the ground truth, $t_i = \{\delta \hat{c}_i,\delta \hat{l}_i\}$ represents predicted relative offset to anchor segments or proposals.
$t_i^* = \{\delta c_i,\delta l_i \}$ represents the coordinate transformation of ground truth segments to anchor segments or proposals. The coordinate transformations are computed as follows:
\begin{equation}
\vspace{-0.1in}
\left\{
\begin{gathered}
  \delta c_i = (c_i^* - c_i) / l_i \\
  \delta l_i = log(l_i^* / l_i)
\end{gathered}
\right.
\label{eq:transformation}
\end{equation}
where $c_i$ and $l_i$ are the center location and the length of anchor segments or proposals while $c_i^*$ and $l_i^*$ denote the same for the ground truth activity segments.

In our \modelname model, the above loss function is applied for both the temporal proposal subnet and the activity classification subnet.
In the proposal subnet, the binary classification loss $L_{cls}$ predicts whether the proposal contains an activity or not, and the regression loss $L_{reg}$ optimizes the relative displacement between proposals and ground truths.
In the proposal subnet the losses are activity class agnostic.
For the activity classification subnet, the multiclass classification loss $L_{cls}$ predicts the specific activity class for the proposal, and the number of classes are the number of activities plus one for the background.
The regression loss $L_{reg}$ optimizes the relative displacement between activities and ground truths.
All four losses for the two subnets are optimized jointly.

\subsection{Prediction}
\label{sec:prediction}
Activity prediction in R-C3D consists of two steps.
First, the proposal subnet generates candidate proposals and predicts the start-end time offsets as well as proposal score for each.
Then the proposals are refined via NMS with threshold value 0.7.
After NMS, the selected proposals are fed to the classification network to be classified into specific activity classes, and the activity boundaries of the predicted proposals are further refined by the regression layer.
The boundary prediction in both proposal subnet and classification subnet is in the form of relative displacement of center point and length of segments.
In order to get the start time and end time of the predicted proposals or activities, inverse coordinate transformation to Equation \ref{eq:transformation} is performed.

\modelname accepts variable length input videos.
However, to take advantage of the vectorized implementation in fast deep learning libraries, we pad the last few frames of short videos with last frame, and break long videos into buffers (limited by memory only).
NMS at a lower threshold (0.1 less than the mAP evaluation threshold) is applied to the predicted activities to get the final activity predictions.

\section{Experiments}
We evaluate \modelname on three large-scale activity detection datasets - THUMOS'14~\cite{THUMOS14}, Charades~\cite{sigurdsson2016hollywood} and ActivityNet~\cite{caba2015activitynet}.
Sections~\ref{exp:thumos14},~\ref{exp:activitynet},~\ref{exp:charades} provide the experimental details and evaluation results on these three datasets.
Results are shown in terms of mean Average Precision - mAP@$\alpha$ where $\alpha$ denotes different Intersection over Union (IoU) thresholds, as is the common practice in the literature.
Section~\ref{exp:speed} provides the detection speed comparison with state-of-the-art methods.

\subsection{Experiments on THUMOS'14}
\label{exp:thumos14}

\begin{table}[!t]
\centering
\caption{Activity detection results on THUMOS'14 (in percentage). mAP at different IoU thresholds $\alpha$ are reported. The top three performers on the THUMOS'14 challenge leaderboard and other results reported in existing papers are shown.}
\small
\begin{tabular}{l || c c c c c} 
\hline
 ~ & \multicolumn{5}{c}{$\alpha$} \\
 ~ & \!\!0.1  & \!\!0.2  & \!\!0.3  & \!\!0.4 & \!\!0.5 \\ \hline
 \!\!Karaman et al.~\cite{karaman2014fast} & \!\!4.6  & \!\!3.4  &  \!\!2.1 & \!\!1.4 & \!\!0.9 \\ 
 \!\!Wang et al.~\cite{wang2014action} & \!\!18.2  & \!\!17.0  & \!\!14.0 & \!\!11.7 &  \!\!8.3 \\ 
 \!\!Oneata et al.~\cite{oneata2014lear} & \!\!36.6  &  \!\!33.6 & \!\!27.0  & \!\!20.8 & \!\!14.4 \\ 
 \!\!Heilbron et al.~\cite{caba2016fast} & \!\!- & \!\!-  & \!\!-  & \!\!- & \!\!13.5 \\ 
 \!\!Escorcia et al.~\cite{escorcia2016daps} & \!\!- & \!\!-  & \!\!-  & \!\!- & \!\!13.9 \\ 
 \!\!Richard et al.~\cite{Richard_2016_CVPR} & \!\!39.7 & \!\!35.7  & \!\!30.0  & \!\!23.2 & \!\!15.2 \\ 
 \!\!Yeung et al.~\cite{yeung2016end} & \!\!48.9 &  \!\!44.0 &  \!\!36.0 & \!\!26.4 & \!\!17.1 \\ 
 \!\!Yuan et al.~\cite{yuan2016temporal} & \!\!51.4 & \!\!42.6  &  \!\!33.6 & \!\!26.1 & \!\!18.8 \\ 
 \!\!Shou et al.~\cite{shou2016temporal} & \!\!47.7 & \!\!43.5  & \!\!36.3  & \!\!28.7 & \!\!19.0 \\ 
 \!\!Shou et al.~\cite{shou2017cdc} & \!\!- & \!\!-  & \!\!40.1  & \!\!29.4 & \!\!23.3 \\ \hline
\!\!R-C3D (our one-way buffer) \!\!\!& \!\!51.6 & \!\!49.2 & \!\!42.8  & \!\!33.4 & \!\!27.0\\
\!\!R-C3D (our two-way buffer) \!\!\!& \!\!\textbf{54.5} & \!\!\textbf{51.5}  & \!\!\textbf{44.8}  & \!\!\textbf{35.6} & \!\!\textbf{28.9} \\ \hline 
\end{tabular}
\label{tab:res_thumos14}
\vspace{-0.2in}
\end{table}

THUMOS'14 activity detection dataset contains over 24 hours of video from 20 different sport activities.
The training set contains 2765 trimmed videos while the validation and the test sets contain 200 and 213 untrimmed videos respectively.
This dataset is particularly challenging as it consists of very long videos (up to a few hundreds of seconds) with multiple activity instances of very small duration (up to few tens of seconds).
Most videos contain multiple activity instances of the same activity class.
In addition, some videos contain activity segments from different classes.

\noindent{\textbf{Experimental Setup}:}
We divide 200 untrimmed videos from the validation set into 180 training and 20 held out videos to get the best hyperparameter setting.
All 200 videos are used as the training set and the final results are reported on 213 test videos.
Since the GPU memory is limited, we first create a buffer of 768 frames at 25 frames per second (fps) which means approximately 30 seconds of video.
Our choice is motivated by the fact that 99.5\% of all activity segments in the validation set (used here as the training set) are less than 30 seconds long.
These buffers of frames act as inputs to \modelname.
We can create the buffer by sliding from the beginning of the video to the end, denoted as the `one-way buffer'.
An additional pass from the end of the video to the beginning is used to increase the amount of training data, denoted as the `two-way buffer'.
We initialize the 3D ConvNet part of our model with C3D weights trained on Sports-1M and finetuned on UCF101 released by the authors in~\cite{tran2015learning}.
We allow all the layers of \modelname to be trained on THUMOS'14 with a fixed learning rate of 0.0001.

\begin{table}[!t]
\centering
\caption{Per-class AP at IoU threshold $\alpha=0.5$ on THUMOS'14 (in percentage).}
\small
 \begin{tabular}{l || c c c c} 
 \hline
 ~ & \cite{oneata2014lear} & \cite{yeung2016end} &  \cite{shou2016temporal} & R-C3D (ours)\\ \hline
 Baseball Pitch  & 8.6 & 14.6 & 14.9&  \bf{26.1}  \\ 
 Basketball Dunk & 1.0 & 6.3 &20.1 & \bf{54.0}  \\ 
 Billiards & 2.6 & \bf{9.4} & 7.6 & 8.3  \\ 
 Clean and Jerk  & 13.3 & \bf{42.8} &24.8 & 27.9  \\ 
 Cliff Diving    & 17.7 & 15.6 &27.5 & \bf{49.2}  \\ 
 Cricket Bowling & 9.5 & 10.8 & 15.7& \bf{30.6}  \\ 
 Cricket Shot    & 2.6 & 3.5 &\bf{13.8} & 10.9  \\ 
 Diving & 4.6    & 10.8 & 17.6 & \bf{26.2}  \\ 
 Frisbee Catch   & 1.2 & 10.4 &15.3 & \bf{20.1}  \\ 
 Golf Swing      & \bf{22.6}  & 13.8 &18.2 & 16.1  \\ 
 Hammer Throw    & 34.7 & 28.9 &19.1 & \bf{43.2}  \\ 
 High Jump       & 17.6 & \bf{33.3} &20.0 & 30.9 \\ 
 Javelin Throw   & 22.0 & 20.4 &18.2 & \bf{47.0}  \\ 
 Long Jump       & 47.6 & 39.0 &34.8 & \bf{57.4}  \\ 
 Pole Vault      & 19.6 & 16.3 &32.1 & \bf{42.7}  \\ 
 Shotput & 11.9  & 16.6 &12.1 & \bf{19.4}  \\ 
 Soccer Penalty  & 8.7 & 8.3 &\bf{19.2} & 15.8  \\ 
 Tennis Swing    & 3.0 & 5.6 &\bf{19.3} & 16.6  \\ 
 Throw Discus    & \bf{36.2} & 29.5 &24.4 & 29.2 \\ 
 Volleyball Spiking & 1.4 & 5.2 &4.6 & \bf{5.6} \\ \hline

 mAP@0.5 &14.4 & 17.1 & 19.0 & \bf{28.9} \\ \hline 
 \end{tabular}
\label{tab:per_class_ap}
\vspace{-0.2in}
\end{table}

The number of anchor segments $K$ chosen for this dataset is 10 with specific scale values [2, 4, 5, 6, 8, 9, 10, 12, 14, 16].
The values are chosen according to the distribution of the activity durations in the training set.
At 25 fps and temporal pooling factor of 8 ($C_{tpn}$ downsamples the input by 8 temporally), the anchor segments correspond to segments of duration  between 0.64 and 5.12 seconds\footnote{$2*8/25=0.64$ and $16*8/25=5.12$}.
Note that, the predicted proposals or activities are relative to the anchor segments but not limited to the anchor boundaries, enabling our model to detect variable-length activities.

\noindent{\textbf{Results}:}
As a sanity check, we first evaluate the performance of the temporal proposal subnet.
A predicted proposal is marked correct if its IoU with a ground truth activity is more than 0.7, otherwise it is considered incorrect.
With this binary setting, precision and recall values of the temporal proposal subnet are 85\% and 83\% respectively.

In Table~\ref{tab:res_thumos14}, we present a comparative evaluation of the activity detection performance of \modelname with existing state-of-the-art approaches in terms of mAP at IoU thresholds 0.1-0.5 (denoted as $\alpha$).
For both the one-way buffer setting and the two-way buffer setting we achieve new state-of-the-art for all five $\alpha$ values.
In the one-way setting, mAP@0.5 is 27.0\% which is an 3.7\% absolute improvement from the state-of-the-art.
The two-way buffer setting further increases the mAP values at all the IoU thresholds with mAP@0.5 reaching as far as 28.9\%.
Our model comprehensively outperforms the current state-of-the-art by a large margin (28.9\% compared to 23.3\% as reported in~\cite{shou2017cdc}).

The Average Precision (AP) for each class in THUMOS'14 at IoU threshold 0.5 for the two-way buffer setting is shown in Table~\ref{tab:per_class_ap}.
\modelname outperforms the all the methods in most classes and shows significant improvement (by more than 20\% absolute AP over the next best) for activities \textit{e.g.}, Basketball Dunk, Cliff Diving, and Javelin Throw.
For some of the activities, our method is only second to the best performing ones by a very small margin (\textit{e.g.}, Billiards or Cricket Shot).
Figure~\ref{fig:vis_thumos14} shows some representative qualitative results from two videos in this dataset.

\subsection{Experiments on ActivityNet}
\label{exp:activitynet}

The ActivityNet~\cite{caba2015activitynet} dataset consists of untrimmed videos and is released in three versions.
We use the latest release (1.3) which has 10024, 4926 and 5044 videos containing 200 different types of activities in the train, validation and test sets respectively.
Most videos contain activity instances of a single class covering a great deal of the video.
Compared to THUMOS'14, this is a large-scale dataset both in terms of the number of activities involved and the amount of video.
Researchers have taken part in the ActivityNet challenge~\cite{activitynetChallenge} held on this dataset.
The performances of the participating teams are evaluated on test videos for which the ground truth annotations are not public.
In addition to evaluating on the validation set, we show our performance on the test set after evaluating it on the challenge server.

\noindent{\textbf{Experimental Setup}:}
Similar to THUMOS'14, the length of the input buffer is set to 768 but, as the videos are long, we sample frames at 3 fps to fit it in the GPU memory.
This makes the duration of the buffer approximately 256 seconds covering over 99.99\% training activities.
The considerably long activity durations prompt us to set the number of anchor segments $K$ to be as high as 20.
Specifically, we chose the following scales - [1, 2, 3, 4, 5, 6, 7, 8, 10, 12, 14, 16, 20, 24, 28, 32, 40, 48, 56, 64].
Thus the shortest and the longest anchor segments are of durations 2.7 and 170 seconds respectively covering 95.6\% training activities.

Considering the vast domain difference of the activities between Sports-1M and ActivityNet, we finetune the Sports-1M pretrained 3D ConvNet model~\cite{tran2015learning} with the training videos of ActivityNet.
We initialize the 3D ConvNet with these finetuned weights.
AcitivityNet being a large scale dataset, the training takes more epochs.
As a speed-efficiency trade-off, we freeze the first two convolutional layers in our model during training.
The learning rate is kept fixed at $10^{-4}$ for first 10 epochs and is decreased to $10^{-5}$ for the last 5 epochs.
Based on the improved results on the THUMOS'14, we choose the two-way buffer setting with horizontal flipping of frames for data augmentation.

\begin{table}[!t]
\centering
\caption{Detection results on ActivityNet in terms of mAP@0.5 (in percentage). The top half of the table shows performance from methods using additional handcrafted features while the bottom half shows approaches using deep features only (including ours). Results for~\cite{Singh2016a} are taken from~\cite{activitynetChallenge}}
\small
\begin{tabular}{l || c |c c c c} 
\hline
 ~ & train data & validation  & test   \\ \hline
G. Singh \textit{et. al}.~\cite{Singh2016b} & train & 34.5  & 36.4   \\
B. Singh  \textit{et. al.}~\cite{Singh2016a} & train+val & -  & 28.8   \\ \hline
UPC ~\cite{montes2016temporal} & train & 22.5  & 22.3   \\
\modelname (ours) & train & \textbf{26.8} & \textbf{26.8}\\ 
\modelname (ours) & train+val & - & \textbf{28.4} \\ \hline
\end{tabular}
\label{res:activitynet}
\end{table}

\noindent{\textbf{Results}:}
In Table~\ref{res:activitynet} we show the performance of \modelname and compare with existing published approaches.
Results are shown for two different settings.
In the first setting, only the training set is used for training and the performance is shown for either the validation or test data or both.
In the second setting, training is done on both training and validation sets while the performance is shown on the test set.
The table shows that the proposed method does achieve a performance better than methods not using handcrafted features \textit{e.g.}, UPC~\cite{montes2016temporal}.
UPC is the most fair comparison as it also uses only C3D features.
However, it relies on a strong assumption that each video in ActivityNet just contains one activity class.
Our approach obtains an improvement of 4.3\% on the validation set and 4.5\% on the test set over UPC~\cite{montes2016temporal} in terms of mAP@0.5 without any such strong assumptions.
When both training and validation sets are used for training, the performance improves further by 1.6\%. 
The ActivityNet Challenge in 2017 introduced a new evaluation metric where mAP at 10 evenly distributed thresholds between 0.5 and 0.95 are averaged to get the \emph{average mAP}.
Using only training data to train R-C3D, the average mAP for the validation and test set are 12.7\% and 13.1\% respectively.
On the other hand, if both training and validation data is used during training, the average mAP for the test set increases to 16.7\% showing the benefit of our end-to-end model when more data is available for training.

\begin{figure*}[!t]
\centering
\subfigure[THUMOS'14]{
\includegraphics[width=0.9\linewidth]{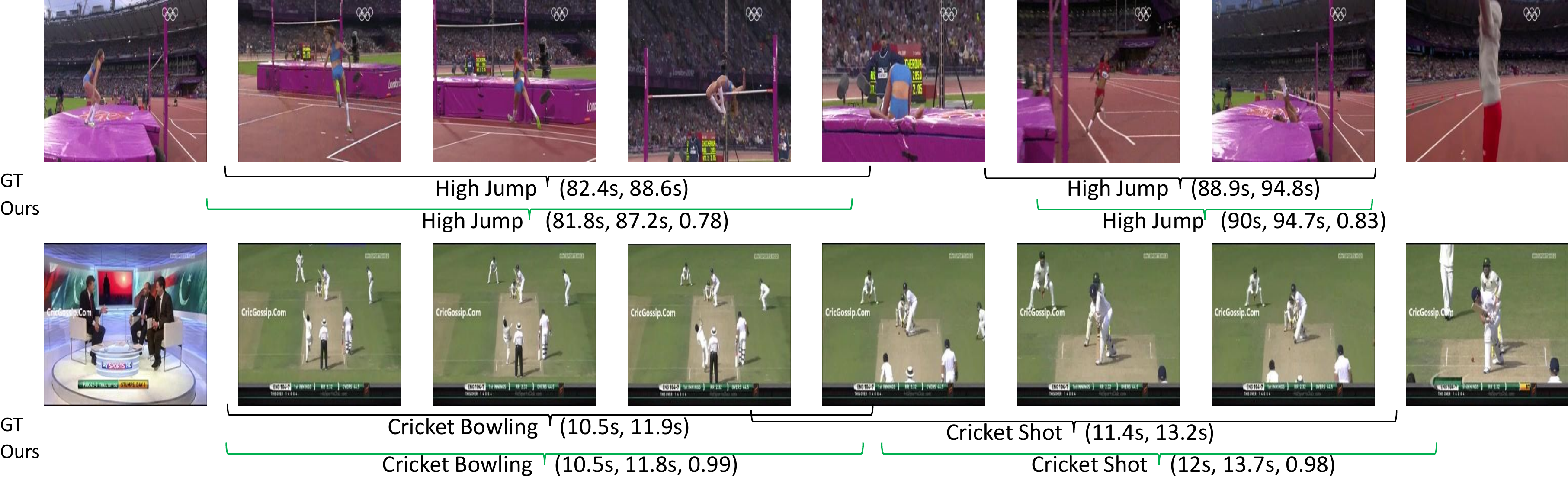}
\label{fig:vis_thumos14}
}
\subfigure[ActivityNet]{
\label{fig:vis_activitynet}
\includegraphics[width=0.9\linewidth]{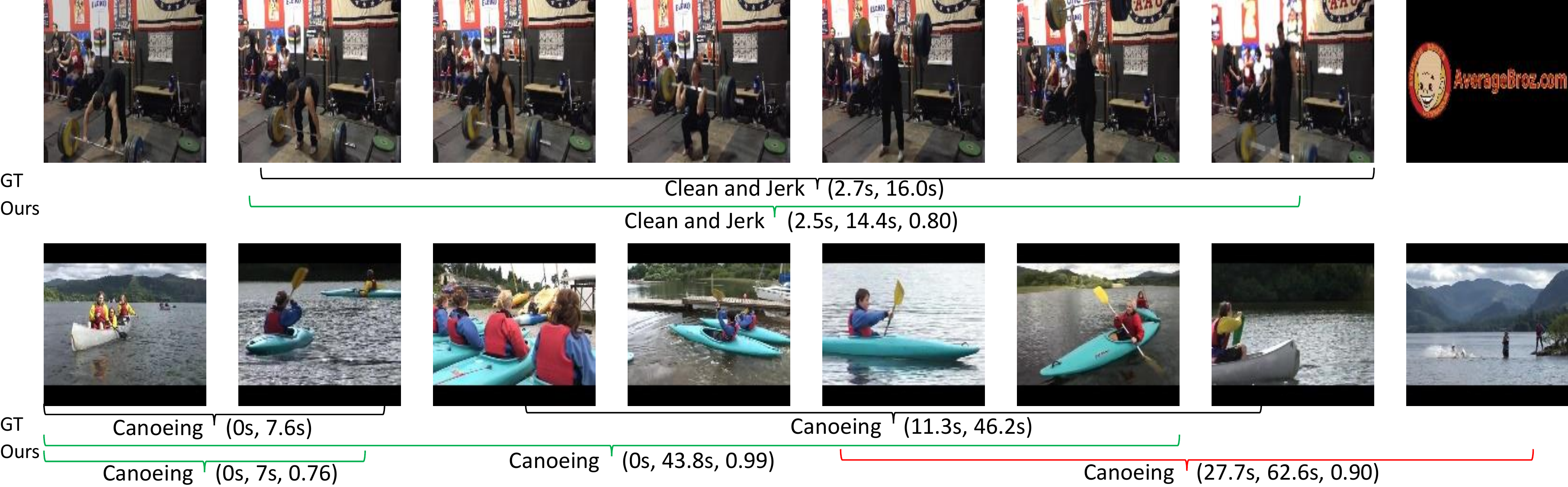}}
\subfigure[Charades]{
\label{fig:vis_charades}
\includegraphics[width=0.9\linewidth]{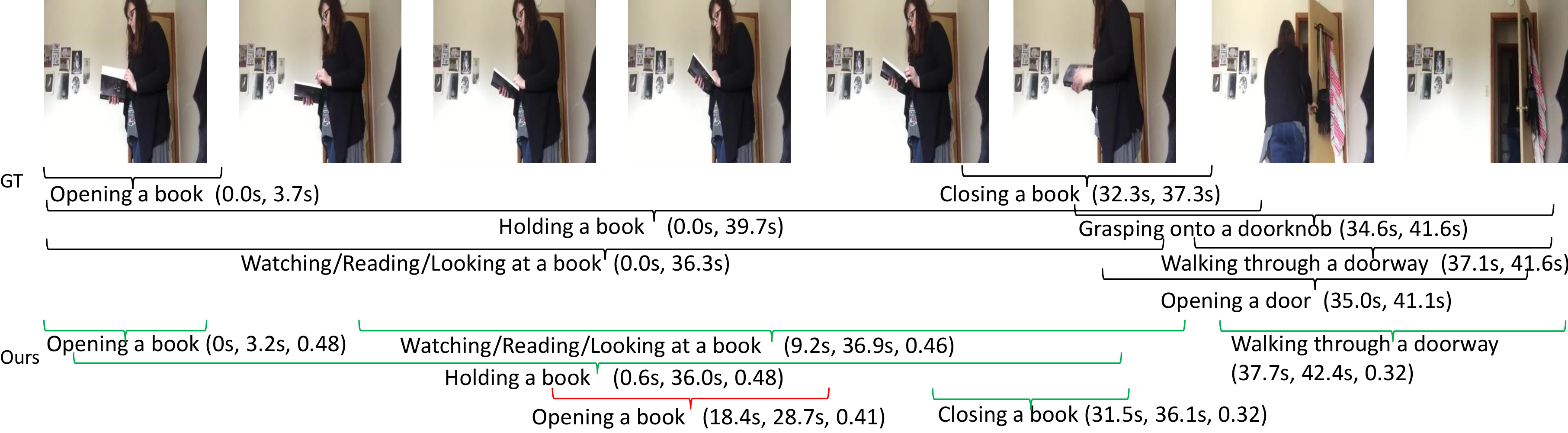}}
\caption
{Qualitative visualization of the predicted activities by \modelname (best viewed in color). Figure \subref{fig:vis_thumos14} and \subref{fig:vis_activitynet} show results for two videos each in THUMOS'14 and ActivityNet.~\subref{fig:vis_charades} shows the result for one video from Charades. Groundtruth activity segments are marked in black. Predicted activity segments are marked in green for correct predictions and in red for wrong ones. Predicted activities with IoU more than 0.5 are considered as correct. Corresponding start-end times and confidence score are shown inside brackets.}
\vspace{-0.2in}
\label{fig:qualitative}
\end{figure*}

\modelname falls slightly behind~\cite{Singh2016a} which uses LSTM based tracking and performs activity prediction using deep features as well as optical flow features from the tracked trajectories.
The approach in~\cite{Singh2016b} also uses handcrafted motion features like MBH on top of inception and C3D features in addition to dynamic programing based post processing.
However, the heavy use of an ensemble of hand-engineered features and dataset dependent heuristics not only stops these methods from learning in an end-to-end fashion but makes them less general across datasets.
Unlike these methods, \modelname is trainable completely end-to-end and is easily extensible to other datasets with little parameter tuning, providing better generalization performance.
Our method is also capable of using hand engineered features with a possible boost to performance, and we keep this as a future task.
Figure~\ref{fig:vis_activitynet} shows some representative qualitative results from this dataset.

\subsection{Experiments on Charades}
\label{exp:charades}

Charades~\cite{sigurdsson2016hollywood} is a recently introduced dataset for activity classification and detection.
The activity detection task involves daily life activities from 157 classes.
The dataset consists of 7985 train and 1863 test videos.
The videos are recorded by Amazon Mechanical Turk users based on provided scripts.
Apart from low illumination, diversity and casual nature of the videos containing day-to-day activities, an additional challenge of this dataset is the abundance of overlapping activities, sometimes multiple activities having exactly the same start and end times (typical examples include pairs of activities like `holding a phone' and `playing with a phone' or `holding a towel' and `tidying up a towel').

\noindent{\textbf{Experimental Setup}:}
For this dataset we sample frames at 5 fps, and the input buffer is set to contain 768 frames.
This makes the duration of the buffer approximately 154 seconds covering all the ground truth activity segments in Charades train set.
As the activity segments for this dataset are longer, we choose the number of anchor segments $K$ to be 18 with specific scale values [1, 2, 3, 4, 5, 6, 7, 8, 10, 12, 14, 16, 20, 24, 28, 32, 40, 48].
So the shortest anchor segment has a duration of 1.6 seconds and the longest anchor segment has a duration of 76.8 seconds. 
Over 99.96\% of the activities in the training set is under 76.8 seconds.
For this dataset we, additionally, explored slightly different settings of the anchor segment scales, but found that our model is not very sensitive to this hyperparameter. 

We first finetune the Sports-1M pretrained C3D model~\cite{tran2015learning} on the Charades training set at the same 5 fps and initialize the 3D ConvNet part of our model with these finetuned weights.
Next, we train \modelname end-to-end on Charades by freezing the first two convolutional layers in order to accelerate training.
The learning rate is kept fixed at 0.0001 for the first 10 epochs and then decreased to 0.00001 for 5 further epochs.
We augment the data by following the two-way buffer setting and horizontal flipping of frames.

\begin{table}[!t]
\centering
\caption{Activity detection results on Charades (in percentage). We report the results using the same evaluation metric as in~\cite{sigurdsson2016asynchronous}.
}
\small
\begin{tabular}{l || c c } 
\hline
 ~ & \multicolumn{2}{c}{mAP} \\ 
 ~ & \!\!\! standard \!\!\!\!\!\! & post-process \\ \hline
 
\!\!\!Random ~\cite{sigurdsson2016asynchronous} & 4.2 & 4.2 \\ 
\!\!\!RGB ~\cite{sigurdsson2016asynchronous} & 7.7 & 8.8 \\ 
\!\!\!Two-Stream ~\cite{sigurdsson2016asynchronous} & 7.7 & 10.0 \\ 
\!\!\!Two-Stream+LSTM ~\cite{sigurdsson2016asynchronous} \!\!\! & 8.3 &  8.8 \\ 
\!\!\!Sigurdsson et al.~\cite{sigurdsson2016asynchronous} & 9.6 & 12.1 \\ \hline
\!\!\!\modelname (ours)& \bf{12.4} & \bf{12.7}\\ \hline 
\end{tabular}
\vspace{-0.2in}
\label{res:charades}
\end{table}

\noindent{\textbf{Results}:}
Table~\ref{res:charades} provides a comparative evaluation of the proposed model with various baseline models reported in~\cite{sigurdsson2016asynchronous}.
This approach~\cite{sigurdsson2016asynchronous} trains a CRF based video classification model (asynchronous temporal fields) and evaluates the prediction performance on 25 equidistant frames by making a multi-label prediction for each of these frames.
The activity localization result is reported in terms of mAP metric on these frames.
For a fair comparison, we map our activity segment prediction to 25 equidistant frames and evaluate using the same mAP evaluation metric.
A second evaluation strategy proposed in this work relies on a post-processing stage where the frame level predictions are averaged across 20 frames leading to more spatial consistency.
As shown in the  Table~\ref{res:charades}, our model outperforms the asynchronous temporal fields model proposed in~\cite{sigurdsson2016asynchronous} as well as the different baselines reported in the same paper.
While the improvement over the standard method is as high as 2.8\%, the improvement after the post-processing is not as high.
One possible reason could be that our end-to-end fully convolutional model captures the spatial consistency implicitly without requiring any manually-designed postprocessing.

Following the standard practice we also evaluated our model in terms of mAP@0.5 which comes out to be 9.3\%.
The performance is not at par with other datasets presumably because of the inherent challenges involved in Charades \textit{e.g.}, the low illumination indoor scenes or the multi-label nature of the data.
Initialization with a better C3D classification model trained on indoor videos with these challenging conditions may further boost the performance.
Figure~\ref{fig:vis_charades} shows some representative qualitative results from one video in this dataset.

One of the major challenges of this dataset is the presence of a large number of temporally overlapping activities.
The results show that our model is capable of handling such scenarios.
This is achieved by the ability of the proposal subnet to produce possibly overlapping activity proposals and is further facilitated by region offset regression.

\subsection{Activity Detection Speed}
\label{exp:speed}
\begin{table}[!t]
\centering
\caption{Activity detection speed during inference.}
\small
\begin{tabular}{l || c } 
\hline
~ & FPS   \\ \hline
S-CNN~\cite{shou2016temporal} & 60   \\ 
DAP~\cite{escorcia2016daps} & 134.1   \\ \hline
R-C3D (ours on Titan X Maxwell)& \bf{569}  \\ \hline 
R-C3D (ours on Titan X Pascal)& \bf{1030}  \\ \hline 
\end{tabular}
\vspace{-0.2in}
\label{res:speed}
\end{table}

In this section, we compare detection speed of our model with two other state-of-the-art methods.
The comparison results are shown in Table~\ref{res:speed}.
S-CNN~\cite{shou2016temporal} uses a time-consuming sliding window strategy and predicts at 60 fps.
DAP~\cite{escorcia2016daps} incorporates a proposal prediction step on top of LSTM and predicts at 134.1 fps.
\modelname constructs the proposal and classification pipeline in an end-to-end fashion and these two stages share the features making it significantly faster.
The speed of execution is 569 fps on a single Titan-X (Maxwell) GPU for the proposal and classification stages together.
On the upgraded Titan-X (Pascal) GPU, our inference speed reaches even higher (1030 fps).
One of the reasons of the speedup of \modelname over DAP may come from the fact that the LSTM recurrent architecture in DAP takes time to unroll, while \modelname directly accepts a wide range of frames as input and the convolutional features are shared by the proposal and classification subnets.
\section{Conclusion} 
We introduce R-C3D, the first end-to-end temporal proposal classification network for activity detection.
We evaluate our approach on three large-scale data sets with very diverse characteristics, and demonstrate that it can detect activities faster and more accurately than existing models based on 3D Convnets.
Additional features can be incorporated into \modelname to further boost the activity detection result.
One future direction may be to integrate \modelname with hand-engineered motion features for improved activity prediction without sacrificing speed.

\noindent{\textbf{Acknowledgement}:} We would thank Lu He for the meaningful discussions. This research was supported by the NSF IIS-1212928 grant, the National Geospatial Agency, and a hardware grant from NVIDIA.

{\small
\bibliographystyle{ieee}
\bibliography{latex/egbib}
}

\end{document}